\documentclass[conference]{IEEEtran}
\IEEEoverridecommandlockouts
\def\BibTeX{{\rm B\kern-.05em{\sc i\kern-.025em b}\kern-.08em
    T\kern-.1667em\lower.7ex\hbox{E}\kern-.125emX}}
\usepackage{cite}
\usepackage{amsmath,amssymb,amsfonts}
\usepackage{graphicx}
\usepackage{textcomp}
\usepackage{xcolor}
\usepackage{booktabs}
\usepackage{hyperref}
\usepackage{multirow}
\usepackage{url}
\usepackage{algorithm}
\usepackage{algpseudocode}
\usepackage{amsmath,amsfonts,bm}

\newtheorem{theorem}{Theorem}

\renewcommand\IEEEkeywordsname{Keywords}

\begin{document}

\title{Imputation using training labels and classification via label imputation}

\author{\IEEEauthorblockN{Thu Nguyen}
\IEEEauthorblockA{\textit{Department of Holistic Systems} \\
\textit{Simula Metropolitan}\\
Oslo, Norway}
\and
\IEEEauthorblockN{Tuan L. Vo}
 \IEEEauthorblockA{
\textit{ Department of Computer Science  } \\
 \textit{LCTI, Télécom Paris} \\
 \textit{Institut Polytechnique de Paris} \\
Paris, France}
\and
\IEEEauthorblockN{Pål Halvorsen, Michael Riegler}
\IEEEauthorblockA{\textit{Department of Holistic Systems} \\
\textit{Simula Metropolitan}\\
Oslo, Norway}
}

\maketitle
\begin{abstract} 
Missing data is a common problem in practical data science settings. Various imputation methods have been developed to deal with missing data. However, even though the labels are available in the training data in many situations, the common practice of imputation usually only relies on the input and ignores the label. We propose \textbf{\textit{Classification Based on MissForest Imputation (CBMI)}}, a classification strategy that initializes the predicted test label with missing values and stacks the label with the input for imputation, allowing the label and the input to be imputed simultaneously. In addition, we propose the \textbf{\textit{imputation using labels (IUL)}} algorithm, an imputation strategy that stacks the label into the input and illustrates how it can significantly improve the imputation quality. Experiments show that CBMI has classification accuracy when the test set contains missing data, especially for imbalanced data and categorical data. Moreover, for both the regression and classification, IUL consistently shows significantly better results than imputation based on only the input data.
\end{abstract}

\begin{IEEEkeywords}
classification, missing data, imputation
\end{IEEEkeywords}

\section{Introduction} \label{sec:introduction}
Data can take various forms, including continuous numerical values, discrete values, or a mix of categorical and continuous data. For example, a housing dataset may include continuous features like price and area, along with discrete features such as the number of rooms and bathrooms. Additionally, missing data is a common issue, which can arise from various causes, such as malfunctioning devices, broken sensors, or survey non-responses. Missing data can undermine the integrity of analyses, leading to biased results and incomplete insights. Therefore, addressing missing data is crucial, which has led to the widespread adoption of imputation techniques.

Imputation, as a remedial practice, involves the estimation or prediction of missing values based on observed information. This process becomes particularly intricate when dealing with datasets that encompass diverse data types, including continuous, categorical, and mixed data. In response to this complexity, a spectrum of imputation techniques has been developed, each tailored to handle specific data types and patterns \cite{vu2023conditional,stekhoven2012missforest}. In addition, in recent years, classification or regression models that can handle missing data directly, such as Decision Tree Classifier~\cite{breiman2001random} and LSTM~\cite{ghazi2018robust},  have also captured a lot of attention. However, imputation remained a popular choice as it renders the data complete not only for the classification or regression task but also for data visualization or other inferences~\cite{pham2023correlation}, or improving the accuracy of parameter estimation~\cite{vo2024effects}.

Here, we propose Classification Based on MissForest Imputation (CBMI) to predict the label on testing data without building a classification model on the training set. The method starts by initializing the predicted testing labels with missing values. Then, it stacks the input and the label, the training and testing data together, and uses the missForest algorithm \cite{stekhoven2012missforest} to impute all the missing values in the matrix. The imputation returns a matrix that contains imputed training and testing input, imputed training labels if the training labels contain missing data, and predicted testing labels. Inspired by the idea that leveraging all the available information from the dataset can lead to improved performance~\cite{vo2025dperc}, we also propose \textbf{\textit{imputation using labels (IUL)}}, a framework for imputation that incorporates the labels into the input to enhance imputation quality. However, unlike CBMI, which relies on missForest, IUL can be used with any imputation method. 

In summary, our contributions are as follows:
    (i) We propose the CBMI algorithm for predicting the output of a classification task via imputation, instead of building a model on training data and then predicting on a test set;
    (ii) We propose using training labels to aid the imputation of the training input and illustrate how this can significantly improve the imputation of training input;
    (iii) We conduct various experiments to illustrate that the proposed approaches usually provide better classification accuracy.


\section{Related works}\label{sec:related_work}
A diverse array of techniques, ranging from traditional methods to sophisticated algorithms, has been developed to handle missing data. Each method brings unique strengths, making them suitable for different types of data and analysis scenarios. Traditional approaches, such as complete case analysis, have been widely used due to their simplicity, but they are prone to bias because they exclude observations with missing values. More sophisticated methods, such as multiple imputations \cite{buuren2010mice} or multiple imputations using Deep Denoising Autoencoders \cite{gondara2017multiple}, have emerged to mitigate these challenges. The methods acknowledge the uncertainty associated with the imputation process by generating multiple plausible values for each missing data point. Another notable work is the Conditional Distribution-based Imputation of Missing Values (DIMV) \cite{vu2023conditional} algorithm. The algorithm finds the conditional distribution of features with missing values based on fully observed features, and its imputation step provides coefficients with direct explainability similar to regression coefficients.

Regression and clustering-based methods also play a significant role in addressing missing data. The CBRL and CBRC algorithms \cite{m2020cbrl}, utilizing Bayesian Ridge Regression, showcase the efficacy of regression-based imputation methods. Additionally, the cluster-based local least square method \cite{keerin2013improvement} offers an alternative approach to imputation based on clustering techniques. For big datasets, deep learning imputation techniques are also of great use due to their powerful performance \cite{choudhury2019imputation, mohan2021graphical}.

For continuous data, techniques based on matrix decomposition or matrix completion, such as Polynomial Matrix Completion \cite{fan2020polynomial}, Alternating Least Squares (ALS) \cite{hastie2015matrix}, and Nuclear Norm Minimization \cite{candes2009exact}, have been employed to make continuous data complete, enabling subsequent analysis using regular data analysis procedures.
Various techniques have been explored for categorical data. The use of k-Nearest Neighbors imputation involves identifying missing values, selecting neighbors based on a similarity metric (e.g., Hamming distance), and imputing missing values by a majority vote from the neighbors' known values. Moreover, the missForest imputation method \cite{stekhoven2012missforest} has shown efficacy in handling datasets that comprise a mix of continuous and categorical features. In addition, some other methods that can handle mixed data include 
SICE \cite{khan2020sice}, and HCMM-LD \cite{murray2016multiple}, k-CMM \cite{dinh2021clustering}.

In recent years, some other aspects of imputation, such as scalability, are also being considered. For example, \cite{nguyen2023principal,nguyen2023principalb} use Principle Component Analysis (PCA) to conduct dimension reduction on fully observed features and merge this part with the not fully observed features to impute. This strategy can result in significant improvement in speed. A similar examination is conducted by \cite{nguyen2023faster} where Singular Value Decomposition is used instead of PCA.  Next, \cite{do2023blockwise} propose a Blockwise principal component analysis Imputation for monotone missing data, where Principal Component Analysis (PCA) is conducted on the observed part of each monotone block of the data and uses a chosen imputation technique to impute on merging the obtained principal components. Experiments also show a significant improvement in imputation speed at a minor cost of imputation quality.

In addition, classification methods and models capable of directly handling missing data have been considerably developed, circumventing the need for imputation. These approaches recognize that imputation introduces a layer of uncertainty and potential bias and instead integrates mechanisms to utilize available information effectively. 
Typically tree-based methods that have such capabilities are Gradient-boosted trees \cite{friedman2001greedy}, Decision Tree Classifier \cite{breiman2017classification},  and their implementation are readily available in the Sklearn package \cite{scikit-learn}. 

The Random Forest classifier has made a pivotal contribution to the field of machine learning and is renowned for its robustness and versatility. Originating from ensemble learning principles, Random Forests combine the predictions of multiple decision trees to enhance overall predictive accuracy and mitigate overfitting. This approach was introduced by \cite{breiman2001random}, and since then, it has gained widespread adoption across diverse domains. The classifier's ability to handle high-dimensional data, capture complex relationships, and provide estimates of feature importance has made it particularly valuable. Various extensions and modifications to the original Random Forest algorithm have been proposed to address specific challenges. For instance, methods like Extremely Randomized Trees \cite{geurts2006extremely} introduce additional randomness during the tree-building process, contributing to increased diversity and potentially improved generalization. Additionally, research efforts have explored adapting Random Forests for specialized tasks, such as imbalanced classification, time-series analysis, and feature selection.  
\vspace{-3mm}
\section{Methodology}\label{sec-method}
\subsection{Classification via imputation}
This section details the methodology of the proposed \textbf{Classification Based on MissForest Imputation (CBMI)} method. The algorithm is presented in algorithm \ref{alg-cbmi}. 

Suppose that we have a dataset that consists of the training set $\{\mathbf{X}_{train}, \mathbf{y}_{train}\}$ that has $n_{train}$ samples and the testing set $\{\mathbf{X}_{test}, \mathbf{y}_{test}\}$ that has $n_{test}$ samples. Then, the procedure is as follows:  

At the first step of the algorithm, we stack the training input $\mathbf{X}_{train}$ and the training label $\mathbf{y}_{train}$ in a column-wise manner. Next, in step 2, we initiate the predicted label of the test set $\hat{\mathbf{y}}$ with missing values, denoted by $*$. After that, in step 3, we stack the input of the test set $\mathbf{X}_{test}$ and $\hat{\mathbf{y}}$ in a column-wise manner, and we call this $\mathcal{D}_{test}^*$.

In the fourth step, we stack $\mathcal{D}_{train}$ and $\mathcal{D}_{test}^*$ in a row-wise manner. This gives us $\mathcal{D}^*$. Then, at step 5, we impute $\mathcal{D} ^*$ using the missForest algorithm. This gives us an imputed matrix $\mathcal{D}$ that contains the imputed training, testing input, and the imputed label of the training set if the training data has missing labels and the imputed label $\hat{\mathbf{y}}$ of the test set.  

\begin{algorithm}
\caption{\textbf{CBMI algorithm} }\label{alg-cbmi}
\hspace*{\algorithmicindent} \textbf{Input:} 
\begin{enumerate}
    \item a dataset consists of the training set $\{\mathbf{X}_{train}, \mathbf{y}_{train}\}$ that has $n_{train}$ samples and the testing set $\{\mathbf{X}_{test}, \mathbf{y}_{test}\}$ that has $n_{test}$ samples,
\end{enumerate}

\hspace*{\algorithmicindent} \textbf{Procedure:} 
\begin{algorithmic}[1]
    \State $\mathcal{D}_{train} = [\mathbf{X} _{train}|\mathbf{y}_{train}]$
    \State $\hat{\mathbf{y}} = \begin{pmatrix}
        *\\
        \vdots\\
        *
    \end{pmatrix}$: an NA-initiated vector of size $n_{test}$ where ${n_{test}}$ is the number of sample in the test set
    \State  $\mathcal{D}_{test}^* = [\mathbf{X} _{test}|\hat{\mathbf{y}}]$
    \State  $\mathcal{D}^*=\begin{pmatrix}
        \mathcal{D}_{train} \\
        \mathcal{D}_{test}^*
    \end{pmatrix}$
    \State $\mathcal{D} \leftarrow $ the imputed version of $\mathcal{D} ^*$ using the missForest algorithm
\end{algorithmic}
\textbf{Return:} $\mathcal{D}$, the imputed version of $\hat{\mathbf{y}}$ obtained from $\mathcal{D}$.
\end{algorithm}

So, instead of following the conventional way of imputing the data and then building a model on a training set, CBMI conducts no prior imputation and builds no model on training data to predict the label of the test set. Instead, CBMI combines the training and testing data, including the input and the available and NA-initialized labels, to impute the test set labels.

\subsection{Using labels for imputation}
In this section, we detail the idea of \textbf{\textit{imputation using labels (IUL)}}. For clarity, from now on, we refer to the common imputation procedure that only relies on the input itself, without the labels, as \textbf{\textit{direct imputation (DI)}}.

The IUL process is straightforward and presented in algorithm \ref{alg-iul}. Suppose that we have input data $\mathbf{X}_{train}$ and labels $\mathbf{y}_{train}$.  Then, the procedure starts by stacking the input training data $\mathbf{X}_{train}$ and labels $\mathbf{y}_{train}$ in a column-wise manner in step 1. Then, in step 2, the algorithm uses the imputation method $I$ to impute $\mathcal{D}_{train}$. This gives $\mathcal{D}_{train}^{imp}$. Then in step 3, the algorithm assigns $\mathbf{X}_{train}^{imp}$, the imputed version of $\mathbf{X}_{train}$, to be the submatrix of $\mathcal{D}_{train}$ that consists of all columns except the last column of $\mathcal{D}_{train}$. In addition in step 4, the algorithm assigns $\mathbf{y}_{train}^{imp}$, the imputed version of $\mathbf{y}_{train}$, 
to be the last column of $\mathcal{D}_{train}^{imp}$. The algorithm ends by returning $\mathbf{X}_{train}^{imp}, \mathbf{y}_{train}^{imp}$.

\begin{algorithm}
\caption{\textbf{IUL algorithm} }\label{alg-iul}
\hspace*{\algorithmicindent} \textbf{Input:} 
\begin{enumerate}
    \item input training data $\mathbf{X}_{train}$, labels $\mathbf{y}_{train}$
    \item  imputation algorithm $I$.
\end{enumerate}

\hspace*{\algorithmicindent} \textbf{Procedure:} 
\begin{algorithmic}[1]
    \State $\mathcal{D}_{train} = [\mathbf{X} _{train}|\mathbf{y}_{train}]$
    \State $\mathcal{D}_{train}^{imp} \leftarrow $ the imputed version of $\mathcal{D}_{train}$ using the imputation algorithm $I$.
    \State $\mathbf{X}_{train}^{imp} \leftarrow $ imputed version of $\mathbf{X}_{train}$ that consists of all columns except the last column of $\mathcal{D}_{train}$, 
    \State $\mathbf{y}_{train}^{imp} \leftarrow $ imputed version of $\mathbf{y}_{train}$ if $\mathbf{y}_{train}$ contains missing values. This corresponds to the last column of $\mathcal{D}_{train}^{imp}$.
\end{algorithmic}
\textbf{Return:} $\mathbf{X}_{train}^{imp}, \mathbf{y}_{train}^{imp} $
\end{algorithm}

It is important to note that \textbf{\textit{the IUL algorithm can handle not only $\mathbf{X}_{train}$ with missing data, but also $\mathbf{y}_{train}$ with missing data. }} In case the labels contain missing data, the returned $\mathbf{y}_{train}^{imp}$ is the imputed version of the training labels. So, this also has a great potential to be applied to semi-supervised learning, as in such a setting, the samples without labels can be considered as samples with missing labels. 

Some intuition about why IUL works better in practice can be seen. Firstly, from the feature selection aspect: when we use labels for imputing a feature $f$ in the input, the amount of information increases. Therefore, one can expect the imputation quality to improve unless the feature is not useful or the information from the label that is useful for imputing $f$ can already be retrieved by other feature(s) in the input. For an Ordinary Least Square problem, for example, the MSE is known not to decrease as the number of features increases. However, this may not hold true for other regression or classification models. Secondly, the missing data mechanism is another important factor. For example, multiple imputation is typically effective when data is missing at random (MAR)~\cite{de2013multiple}, where the probability of missingness in a variable depends on other observed variables. By stacking labels to the input, the label feature may reveal hidden dependencies influencing why data is missing, thereby increasing the likelihood that the missing data mechanism is MAR rather than missing completely at random (MCAR). This alignment with MAR makes the imputation process more reliable since multiple imputation (missForest or MICE), performs better under the MAR assumption.

Moreover, there are some reasons for a specific scenario. For example with missForest imputation, which is built on the Random Forest, \textbf{\textit{the difference in the number of choices for building a Random Forest when stacking the label to the input as in IUL can be significantly higher than imputing based only on $\mathbf{X}_{train}$ itself}}. To be more specific, assuming we have $p$ features in $\mathbf{X}_{train}$ and one label $\mathbf{y}$. With IUL, the label can be utilized to aid the imputation of $\mathbf{X}_{train}$, which means that we have $p+1$ features. In particular, if we are to build $m$ model RandomForest and choose $k$ features for each model, then we can have $\begin{pmatrix}
    p+1 \\ k
\end{pmatrix}$ combinations. Meanwhile, {DI} only work on $\mathbf{X}_{train}$ so there are just $\begin{pmatrix}
    p \\ k
\end{pmatrix}$ options. The difference is $
    \begin{pmatrix}
    p+1 \\ k
\end{pmatrix} - \begin{pmatrix}
    p \\ k
\end{pmatrix} = \begin{pmatrix}
    p \\ k-1
\end{pmatrix},
$ which can be extremely large. 

When using the MICE algorithm in the IUL strategy, while the experiments will show that IUL performs competitively compared to DI, IUL is not theoretically guaranteed to outperform DI all the time. The following results illustrate that.

\begin{theorem}\label{theo-1}
Assume that we have a data $\mathcal{D}=(\mathbf{x},\mathbf{z},\mathbf{y})$ of $n$ samples. Here, $\mathbf{x}$ contains missing values, $\mathbf{z}$ is fully observed, and $\mathbf{y}$ is a label feature. For MICE imputation, with the IUL strategy, we construct the model $ \hat{x}=\hat{\gamma}_o+\hat{\gamma}_1z+\hat{\gamma}_2y. $
Meanwhile, DI ignores label feature $\mathbf{y}$ and use the following model 
$  \hat{x}'=\hat{\gamma}_o+\hat{\gamma}_1z. $
Without loss of generality, assume that $y_i \ge 0, i=1,2,...,n$ for label encoding. Let denote 
\begin{align*}
    \mathcal{E}_i &= \hat{\gamma}_2^2(y_i-2\mu_{\mathbf{y}})+2\hat{\gamma}_1\hat{\gamma}_2(z_i-\mu_{\mathbf{z}}), \quad i=1,2,\dots,n.
\end{align*}
Here, the value of $\mathcal{E}_i$ could be negative or non-negative. Thus, we distinguish between two parts by denoting
\begin{align*}
    \mathcal{V}^+ = \sum_{i\in \mathcal{I}^+}y_i\mathcal{E}_i, \text{ where }\mathcal{I}^+ &= \{i : \mathcal{E}_i \ge 0 \}, \\
    \mathcal{V}^- = \sum_{i\in \mathcal{I}^-}y_i\mathcal{E}_i, \text{ where }  \mathcal{I}^- &= \{i : \mathcal{E}_i < 0 \}.
\end{align*}
Then, IUL outperforms DI (i.e., $SSE_{IUL}\le SSE_{DI}$) if and only if $ \mathcal{V}^+ + \mathcal{V}^- \ge 0,$ where $SSE_{IUL}, SSE_{DI}$ are the Sum of Square Error for the model based on IUL and DI, respectively.    
\end{theorem}

The proof of this theorem is provided in Appendix \ref{appendix-proof}.

\section{Experiments}\label{sec-experiments}
\subsection{Experiment setup}
\begin{table}[!ht]
\centering
\caption{Datasets for experiments}\label{tab-datasets}
\vspace*{2mm}
\begin{tabular}{ccc}
\hline
{Datasets} & \# samples & {\# features} \\ \hline
Iris & 150 & 4\\
Liver & 345 & 6 \\ 
Soybean & 47 & 35 \\  
Parkinson & 195 & 22 \\ 
Heart & 267 & 44 \\
Glass & 214 & 9 \\
Car & 1728 & 6 \\
California housing & 20640 & 8 \\
Diabetes & 442 & 10\\
\hline
\end{tabular}
\vspace{-5mm}
\end{table}
\textbf{Datasets and setup:} The datasets for the experiments are from the UCI Machine Learning repository \cite{Dua:2019} and are described in Table~\ref{tab-datasets}. For each dataset, we simulate missing data with missing rates $r$ ranging between 0\% - 80\%. Here, the missing rate is the ratio between the number of missing entries and the total number of entries. After missing data generation, the data is scaled to $[-1,1]$ and impute the data. Each experiment is repeated ten times, and the training-to-testing ratio is 6:4, and we report the mean $\pm$ standard deviation from the runs for accuracy/mean squared error. 

\textbf{Comparing CBMI with IClf and Random Forest:} We compare CBMI with two different algorithms. The first is the two-step approach of \textit{Imputation} using missForest/MICE and then \textit{Classification} using Random Forest Classifier \cite{breiman2001random}, abbreviated as Iclf. The second method uses a random forest to directly classify missing data. The reason for choosing Random Forest as the Classifier is to facilitate fair comparison with CBMI, which relies on Random Forest to impute missing values. The stopping criterion for missForest is the default in the missingpy package\footnote{https://pypi.org/project/missingpy/}. We also conduct the experiments in two settings: the test set is fully observed, and the test set has missing values. In case the test set is fully observed, the missing rate $r$ is for simulating missing values on training data only. In the case where there are missing values in the dataset, the missing values are generated with the same missing rates as in the training input. Moreover, for a fair comparison, when implementing IClf, if the test set contains missing values, then it is merged with the train set for imputation. Note that this is only for the imputation of the test set, i.e., the imputation of the training data is independent of the testing data.

\textbf{Comparing IUL with DI:} In addition, we compare the Imputation Using Labels (IUL) with the Direct Imputation (DI) on the training data approach by simulating missing completely at random data in both the training and testing sets. Next, for the classification datasets, we measure the classification accuracy on the test set using imputed data, as well as the MSE between the imputed and original values on the testing set. Generally, when we achieve better-imputed data, reflected by a lower MSE, we typically expect improved model performance. However, since the input data may also contain inherent noise, MSE alone may not fully capture the model’s effectiveness. Therefore, to provide a more comprehensive evaluation, we also assess performance using accuracy alongside MSE, ensuring that both error minimization and classification success are considered.

\subsection{Results and analysis for CBMI}

\begin{table}[!ht]
\caption{Classification results when \textbf{missing data is present in the test set}. The bold indicates the best performance. In Soybean data, when the missing rate is 80\%, the results are not shown because of disappeared rows.}
\label{tab:test-missing}
\resizebox{\columnwidth}{!}{
\centering
    \begin{tabular}{c@{\hspace{0.12in}}|c@{\hspace{0.12in}}|c@{\hspace{0.12in}}c@{\hspace{0.12in}}c}
    \toprule
    {}&{} & \multicolumn{3}{c}{Accuracy} \\
    Dataset &$r$ &              CBMI (our)&             IClf & RF \\
    \hline
    &20\% & 0.927$\pm$0.03 & \textbf{0.930$\pm$0.031} & 0.909$\pm$0.042 \\
    \multirow{3}{*}{Iris}&40\% & 0.873$\pm$0.044 & \textbf{0.874$\pm$0.038} & 0.830$\pm$0.052 \\
    &60\% & \textbf{0.765$\pm$0.049} & 0.754$\pm$0.060 & 0.685$\pm$0.065 \\
    &80\% & \textbf{0.566$\pm$0.063} & 0.543$\pm$0.070 & 0.500$\pm$0.061 \\
    \hline
    &20\% & 0.644$\pm$0.039 & 0.644$\pm$0.039 & \textbf{0.676$\pm$0.035} \\
    \multirow{3}{*}{Liver}&40\% & 0.596$\pm$0.045 & 0.588$\pm$0.042 & \textbf{0.632$\pm$0.038} \\
    &60\% & 0.551$\pm$0.042 & 0.554$\pm$0.044 & \textbf{0.601$\pm$0.038} \\
    &80\% & 0.539$\pm$0.039 & 0.539$\pm$0.044 & \textbf{0.574$\pm$0.043} \\
    \hline
    \multirow{3}{*}{Soybean} & 20\% & \textbf{0.987$\pm$0.025} & 0.986$\pm$0.031 & 0.972$\pm$0.044 \\
    & 40\% & \textbf{0.935$\pm$0.054} & 0.934$\pm$0.062 & 0.886$\pm$0.100 \\
    &60\% & 0.747$\pm$0.128 & \textbf{0.751$\pm$0.120} & 0.690$\pm$0.145 \\
    \hline
    &20\% & \textbf{0.884$\pm$0.038} & 0.881$\pm$0.039 & 0.872$\pm$0.038 \\
    \multirow{3}{*}{Parkinson}&40\% & \textbf{0.865$\pm$0.044} & 0.863$\pm$0.042 & 0.837$\pm$0.040 \\
    &60\% & \textbf{0.831$\pm$0.036} & 0.820$\pm$0.041 & 0.795$\pm$0.037 \\
    &80\% & \textbf{0.78$\pm$0.044} & 0.778$\pm$0.044 & 0.761$\pm$0.040 \\
    \hline
    &20\% & \textbf{0.811$\pm$0.030} & \textbf{0.811$\pm$0.030} & 0.805$\pm$0.030 \\
    \multirow{3}{*}{Heart}&40\% & 0.798$\pm$0.032 & \textbf{0.801$\pm$0.037} & 0.798$\pm$0.036 \\
    &60\% & 0.794$\pm$0.032 & \textbf{0.798$\pm$0.030} & 0.799$\pm$0.029 \\
    &80\% & 0.772$\pm$0.028 & 0.773$\pm$0.035 & \textbf{0.792$\pm$0.030} \\
    \hline
    &20\% & \textbf{0.712$\pm$0.046} & 0.710$\pm$0.052 & 0.689$\pm$0.047 \\
    \multirow{3}{*}{Glass}&40\% & \textbf{0.634$\pm$0.045} & 0.626$\pm$0.046 & 0.604$\pm$0.053 \\
    &60\% & \textbf{0.535$\pm$0.055} & 0.514$\pm$0.055 & 0.502$\pm$0.058 \\
    &80\% & 0.393$\pm$0.049 & 0.379$\pm$0.049 & \textbf{0.401$\pm$0.049} \\
    \hline
    &20\% & 0.779$\pm$0.015 & 0.773$\pm$0.017 & \textbf{0.818$\pm$0.012} \\
    \multirow{3}{*}{Car}&40\% & \textbf{0.716$\pm$0.018} & 0.703$\pm$0.018 & 0.704$\pm$0.023 \\
    &60\% & \textbf{0.679$\pm$0.015} & \textbf{0.679$\pm$0.026} & 0.630$\pm$0.060 \\
    &80\% & \textbf{0.681$\pm$0.017} & 0.679$\pm$0.038 & 0.584$\pm$0.111 \\
    \bottomrule
    \end{tabular}
    }
\end{table}

In Table~\ref{tab:test-missing}, overall, when missing values are present in the test set, CBMI generally outperforms IClf and Random Forest, except for the Liver dataset. For instance, in the Parkinson dataset, CBMI's accuracy exceeds that of IClf and Random Forest by a range of 1\% to 4\%.

Moreover, note that the number of samples for each class in the Parkinson dataset is (48, 147), while for the Heart dataset, it is (55, 212), for the Glass dataset it is (70, 76, 17, 13, 9, 29), and for the Car dataset, it is (384, 69, 1210, 65), indicating that these are imbalanced datasets. When the test set of these datasets contains missing values, CBMI demonstrates better results than the other algorithms in most cases. For example, in the Glass dataset, as the missing rate increases from 20\% to 80\%, the accuracy gaps between CBMI and IClf are 0.2\%, 0.9\%, 2.1\%, and 1.4\%, respectively.

A similar trend is observed for the categorical datasets, such as Soybean and Car. In the Soybean dataset, CBMI achieves substantially higher accuracy, ranging from nearly 75\% to 99\%, followed by IClf with a similar range, while Random Forest achieves accuracy between 69\% and 97.2\%

\begin{table}
    \caption{Classification results when \textbf{the test input is fully observed}. The bold indicates the best performance. In Soybean data, when the missing rate is 80\%, the results are not shown because of disappeared rows.}
    \label{tab:test-observed}
    \resizebox{\columnwidth}{!}{
    \centering
        \begin{tabular}{c@{\hspace{0.12in}}|c@{\hspace{0.12in}}|c@{\hspace{0.12in}}c@{\hspace{0.12in}}c@{\hspace{0.12in}}}
        \toprule
        {}&{} & \multicolumn{3}{c}{Accuracy} \\
        Dataset &$r$ &              CBMI (our)&             IClf & RF \\
        \hline
        &0\%  & \textbf{0.948$\pm$0.022} & 0.947$\pm$0.023 & 0.947$\pm$0.023 \\
        &20\%  & \textbf{0.948$\pm$0.021} & \textbf{0.948$\pm$0.021} & \textbf{0.948$\pm$0.021} \\
        Iris& 40\%  & 0.949$\pm$0.026 & 0.942$\pm$0.032 & \textbf{0.952$\pm$0.022} \\
        &60\% & 0.942$\pm$0.027 & 0.917$\pm$0.042 & \textbf{0.946$\pm$0.021} \\
        &80\% & 0.933$\pm$0.05 & 0.862$\pm$0.073 & \textbf{0.948$\pm$0.022} \\
        \hline
        
        &0\% & \textbf{0.711$\pm$0.038} & 0.708$\pm$0.036 & 0.708$\pm$0.038 \\
        &20\% & 0.684$\pm$0.037 & 0.676$\pm$0.035 & \textbf{0.706$\pm$0.032} \\
        Liver & 40\% & 0.663$\pm$0.041 & 0.642$\pm$0.042 & \textbf{0.710$\pm$0.032} \\
        &60\% & 0.612$\pm$0.053 & 0.598$\pm$0.043 & \textbf{0.703$\pm$0.034} \\
        &80\% & 0.566$\pm$0.056 & 0.561$\pm$0.056 & \textbf{0.707$\pm$0.035} \\
        
        \hline
        \multirow{4}{*}{Soybean}&0\% & 0.995$\pm$0.016 & 0.995$\pm$0.015 & \textbf{0.997$\pm$0.011} \\
        &20\% & 0.994$\pm$0.021 & 0.989$\pm$0.028 & \textbf{0.995$\pm$0.016} \\
        &40\% & \textbf{0.993$\pm$0.019} & 0.975$\pm$0.063 & \textbf{0.993$\pm$0.019} \\
        &60\% & 0.978$\pm$0.066 & 0.957$\pm$0.077 & \textbf{0.996$\pm$0.013} \\
        \hline
        &0\% & 0.889$\pm$0.039 & 0.887$\pm$0.039 & \textbf{0.891$\pm$0.038} \\
        &20\% & 0.884$\pm$0.037 & 0.884$\pm$0.039 & \textbf{0.887$\pm$0.036} \\
        Parkinson &40\% & 0.883$\pm$0.035 & 0.880$\pm$0.038 & \textbf{0.889$\pm$0.037} \\
        &60\% & 0.867$\pm$0.037 & 0.859$\pm$0.040 & \textbf{0.891$\pm$0.035} \\
        &80\% & 0.813$\pm$0.057 & 0.809$\pm$0.052 & \textbf{0.892$\pm$0.033} \\
        \hline
        &0\% & 0.808$\pm$0.033 & \textbf{0.809$\pm$0.030} & 0.808$\pm$0.032 \\
        &20\% & \textbf{0.809$\pm$0.032} & \textbf{0.809$\pm$0.035} & 0.808$\pm$0.032 \\
        Heart &40\% & \textbf{0.804$\pm$0.033} & 0.802$\pm$0.031 & 0.803$\pm$0.032 \\
        &60\% & 0.797$\pm$0.034 & 0.798$\pm$0.034 & \textbf{0.803$\pm$0.029} \\
        &80\% & 0.782$\pm$0.036 & 0.786$\pm$0.036 & \textbf{0.807$\pm$0.027} \\
        
        \hline
        &0\%& 0.749$\pm$0.049 & \textbf{0.750$\pm$0.048} & 0.749$\pm$0.050 \\
        &20\%& 0.740$\pm$0.043 & 0.730$\pm$0.045 & \textbf{0.749$\pm$0.045} \\
        Glass &40\% & 0.715$\pm$0.050 & 0.690$\pm$0.050 & \textbf{0.746$\pm$0.045} \\
        &60\% & 0.661$\pm$0.057 & 0.622$\pm$0.057 & \textbf{0.751$\pm$0.049} \\
        &80\% & 0.582$\pm$0.070 & 0.506$\pm$0.068 & \textbf{0.753$\pm$0.039} \\
        
        \hline
        & 0 \% & \textbf{0.963$\pm$0.008} & 0.962$\pm$0.009 & 0.961$\pm$0.009 \\
        & 20\% & 0.901$\pm$0.016 & 0.874$\pm$0.017 & \textbf{0.963$\pm$0.009} \\
        Car& 40\% & 0.846$\pm$0.020 & 0.805$\pm$0.018 & \textbf{0.962$\pm$0.010} \\
        &60\% & 0.800$\pm$0.022 & 0.750$\pm$0.022 & \textbf{0.964$\pm$0.010} \\
        &80\% & 0.728$\pm$0.036 & 0.717$\pm$0.023 & \textbf{0.962$\pm$0.010} \\
        \bottomrule
        \end{tabular}
    }
\end{table}

Interestingly, when the test set is fully observed (Table~\ref{tab:test-observed}), Random Forest applied directly to non-imputed (missing) data outperforms the other methods in accuracy. For the Glass and Car datasets, these margins reach nearly 10\% and 20\%, respectively. Comparing CBMI and IClf, CBMI consistently achieves significantly higher accuracy. For instance, in the Car dataset, CBMI outperforms IClf with accuracy gaps of 0.1\%, 2.7\%, 4.1\%, 5\%, and 1.1\%. Notably, the accuracy differences between CBMI and IClf are more pronounced when the test set is fully observed than in the first experiment where missing data is present in the test set.

\subsection{Results and analysis for IUL}
\begin{figure*}[!ht]
    \centering
    \begin{minipage}[b]{0.35\textwidth}
        \centering
        \includegraphics[width=1.05\textwidth]{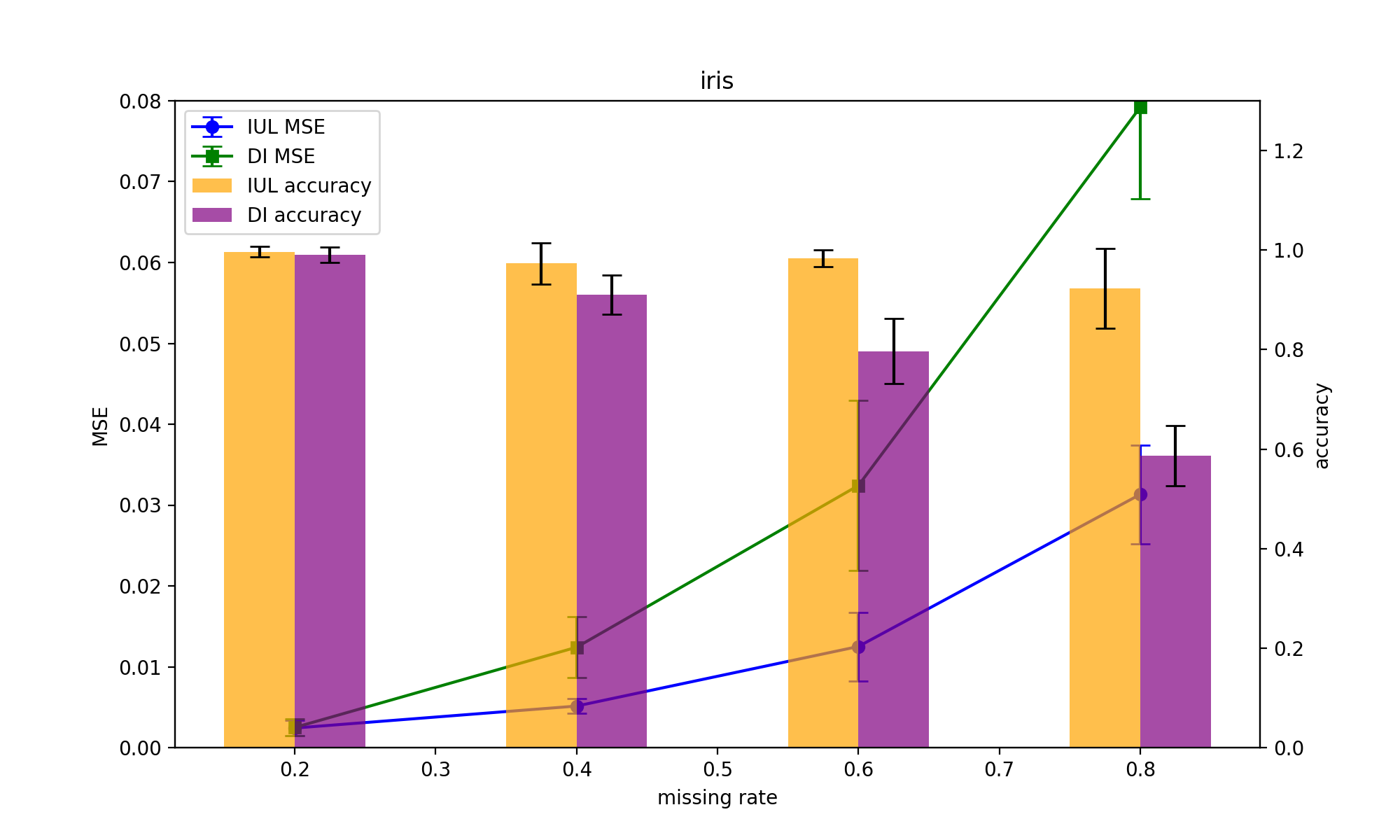}
        \label{fig-iris}
    \end{minipage}
    \\
    \begin{minipage}[b]{0.35\textwidth}
        \centering
        \includegraphics[width=1.05\textwidth]{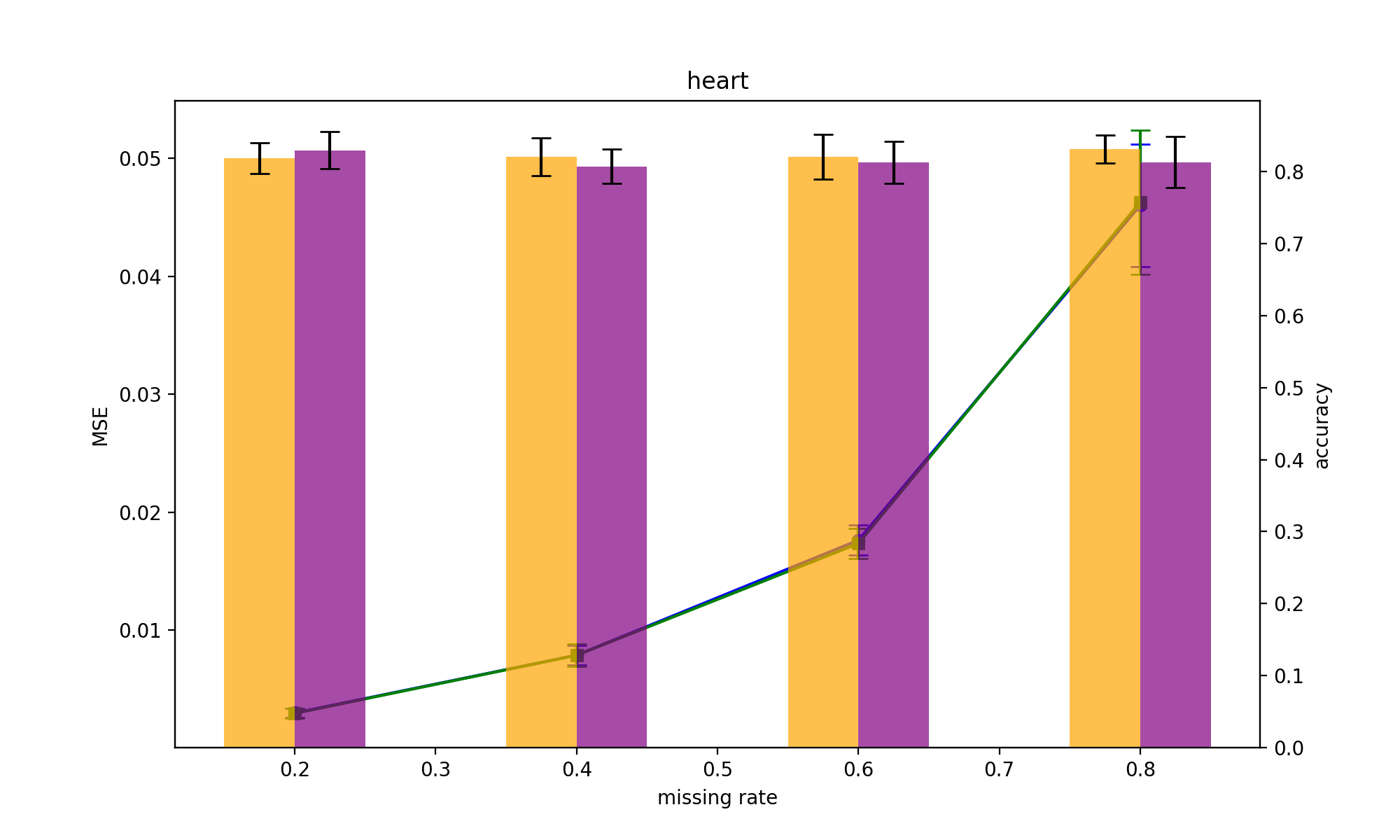}
        \label{fig-heart}
    \end{minipage}
    \begin{minipage}[b]{0.35\textwidth}
        \centering
        \includegraphics[width=1.05\textwidth]{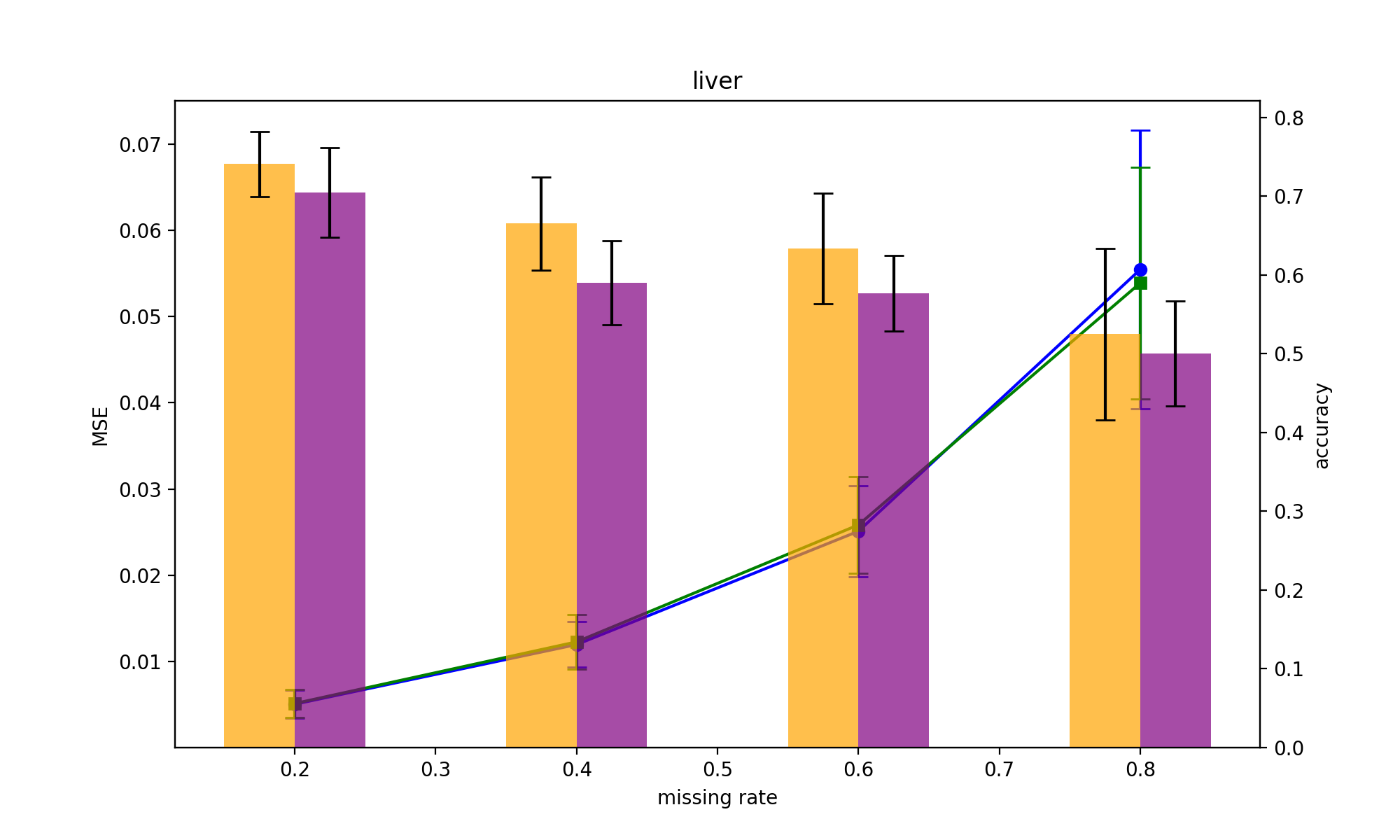}
        \label{fig-liver}
    \end{minipage}
    \begin{minipage}[b]{0.35\textwidth}
        \centering
        \includegraphics[width=1.05\textwidth]{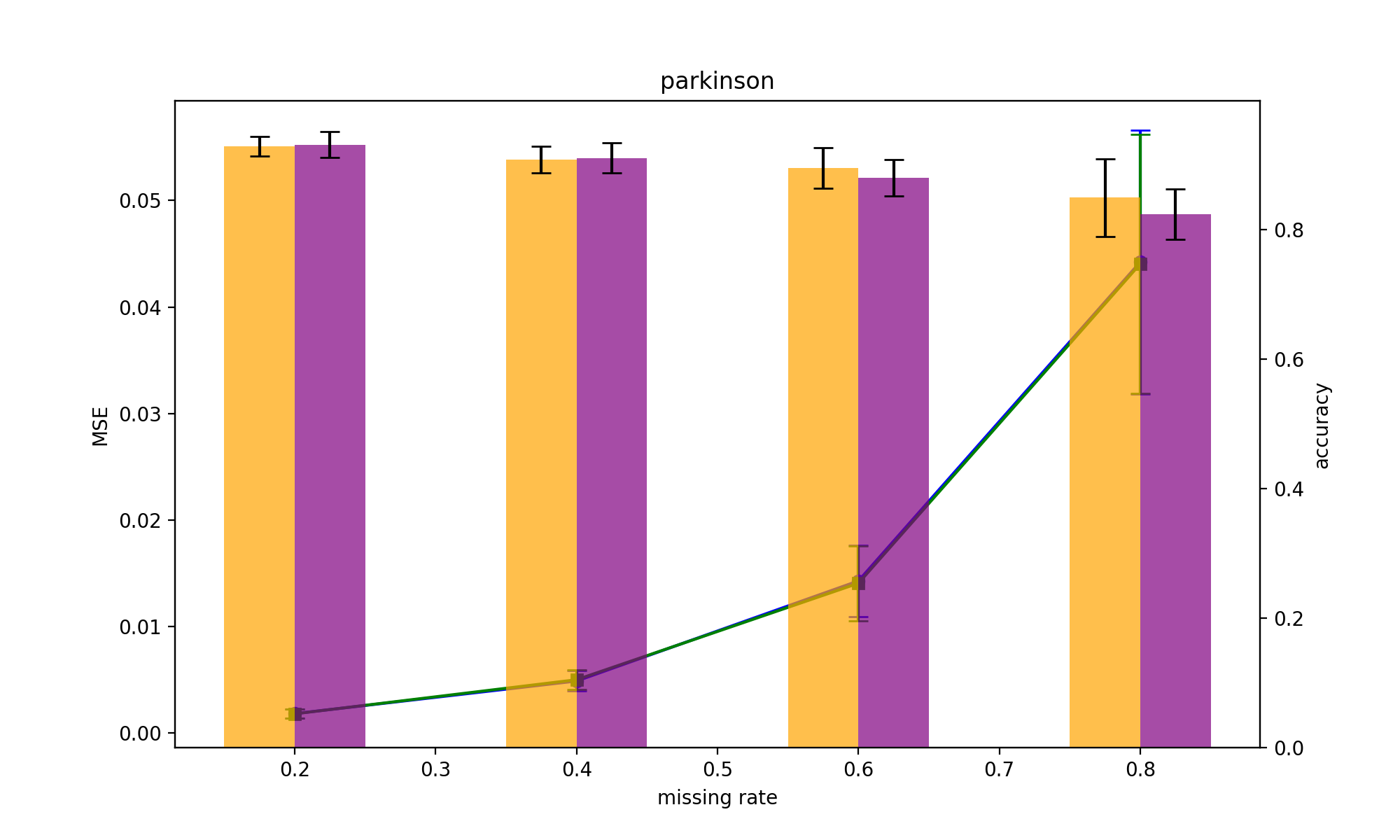}
        \label{fig-parkinson}
    \end{minipage}
    \begin{minipage}[b]{0.35\textwidth}
        \centering
        \includegraphics[width=1.05\textwidth]{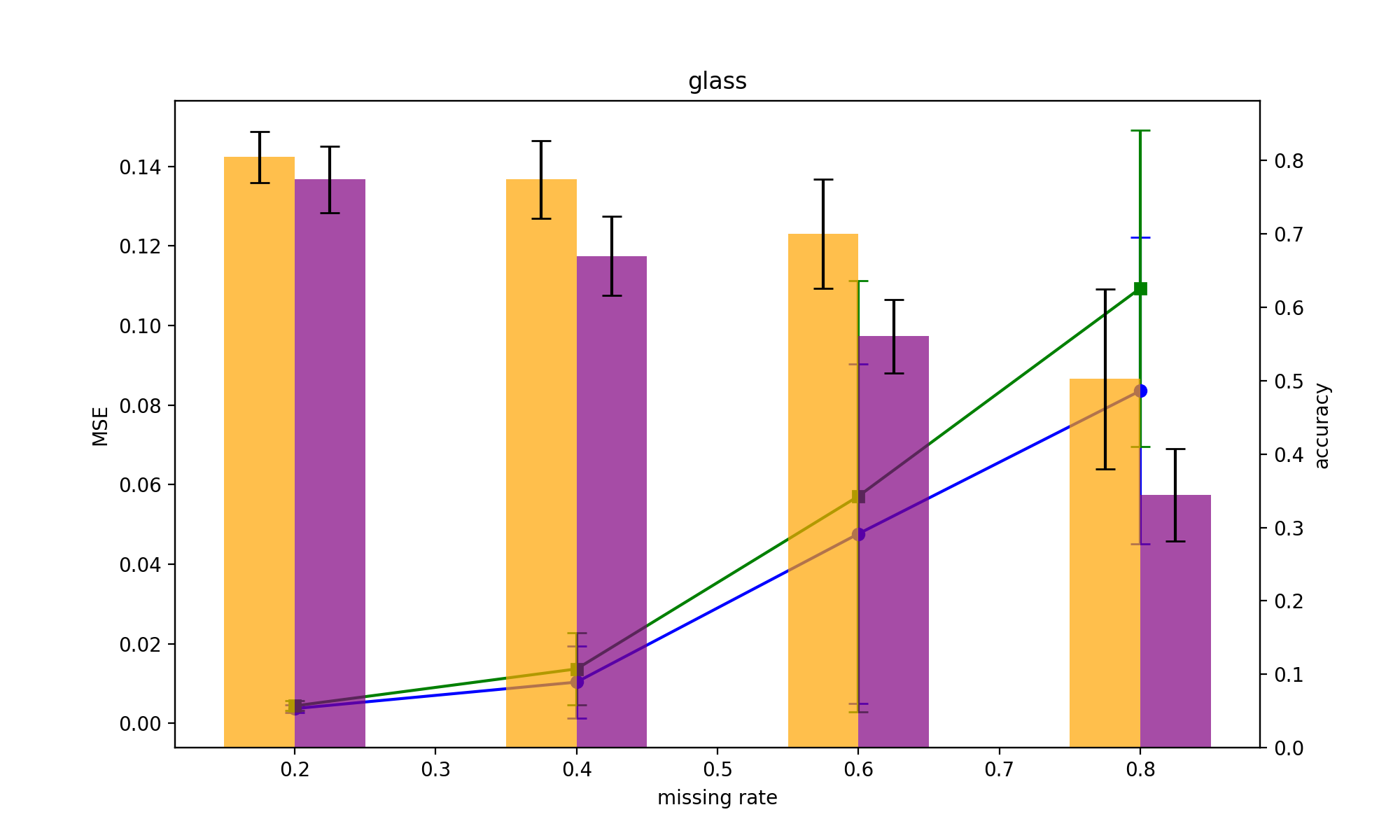}
        \label{fig-glass}
    \end{minipage}
    \caption{Performance of IUL compared to DI with missForest for classification tasks.}
    \label{fig-iul-di}
\end{figure*}

\begin{figure*}[!ht]
    \centering
    \begin{minipage}[b]{0.35\textwidth}
        \centering
        \includegraphics[width=1.05\textwidth]{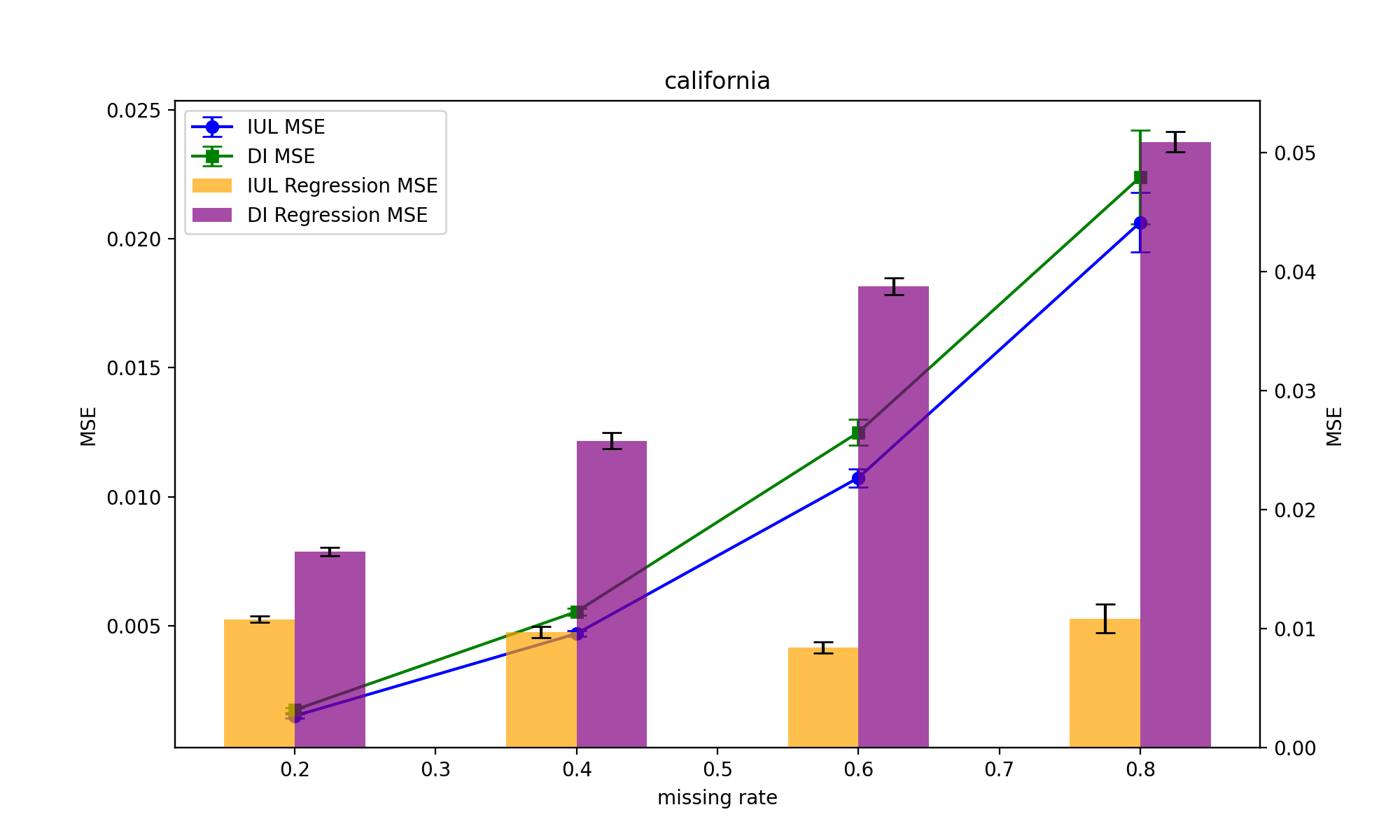}
        \label{fig-cali}
    \end{minipage}
    \begin{minipage}[b]{0.35\textwidth}
        \centering
        \includegraphics[width=1.05\textwidth]{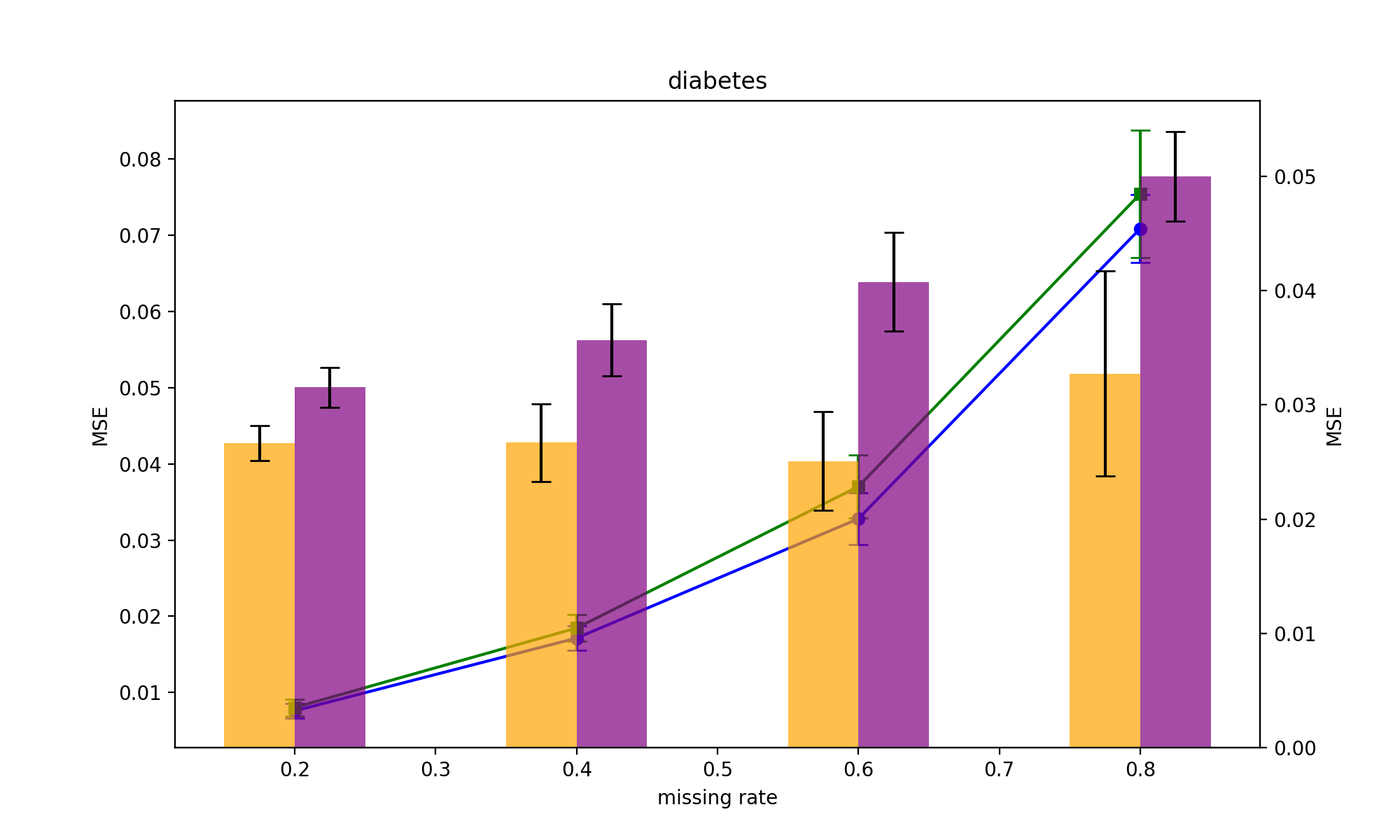}
        \label{fig-diabetes}
    \end{minipage}
    \caption{Performance of IUL compared to DI with missForest for regression tasks.}
    \label{fig-iul-di-regression}
\end{figure*}


\begin{figure*}[!ht]
    \centering
    \begin{minipage}[b]{0.35\textwidth}
        \centering
        \includegraphics[width = 1.05\textwidth]{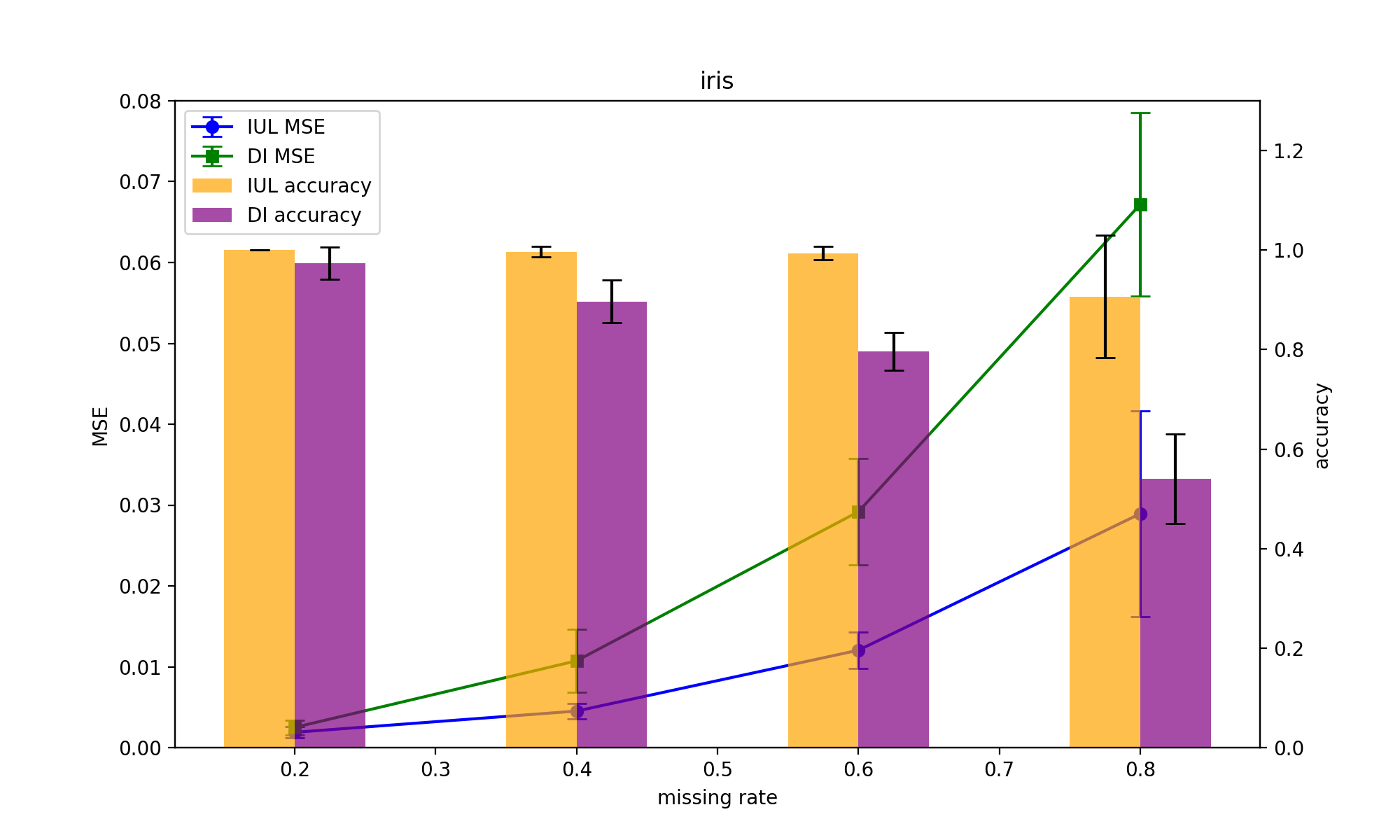}
        \label{fig-iris-mice}
    \end{minipage}   
    \\
    \begin{minipage}[b]{0.35\textwidth}
        \centering
        \includegraphics[width = 1.05\textwidth]{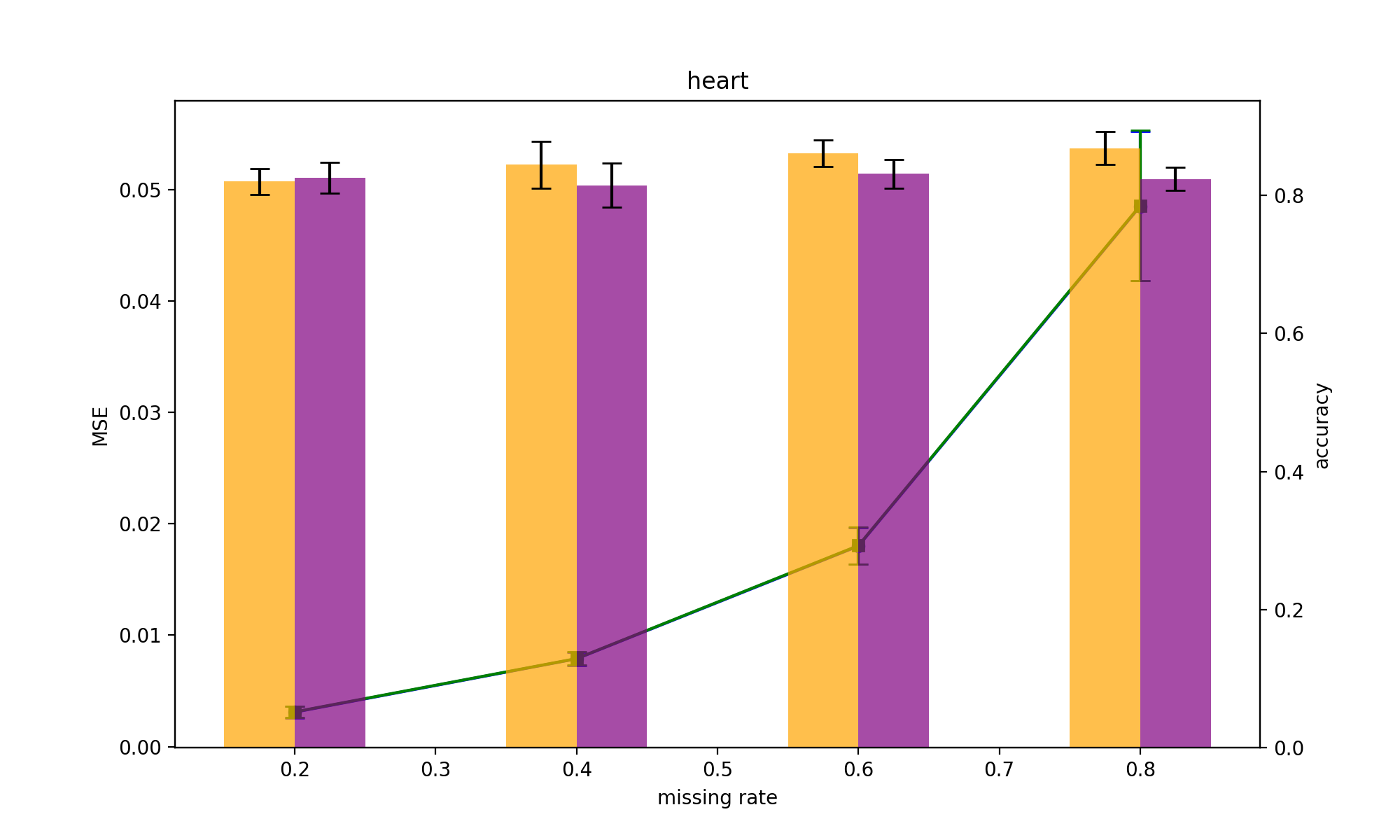}
        \label{fig-heart-mice}
    \end{minipage}
    \begin{minipage}[b]{0.35\textwidth}
        \centering
        \includegraphics[width = 1.05\textwidth]{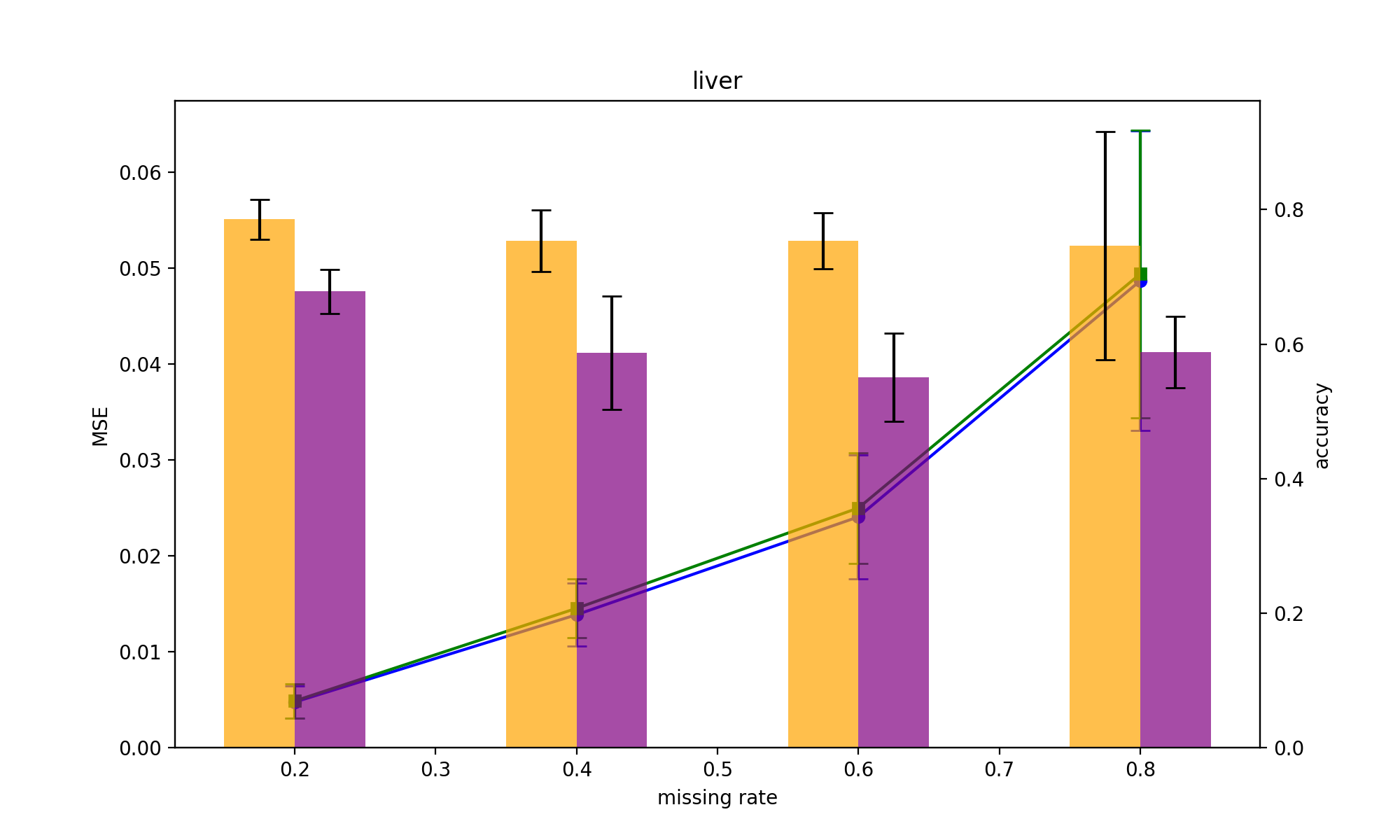}
        \label{fig-liver-mice}
    \end{minipage}
    
    \begin{minipage}[b]{0.35\textwidth}
        \centering
        \includegraphics[width = 1.05\textwidth]{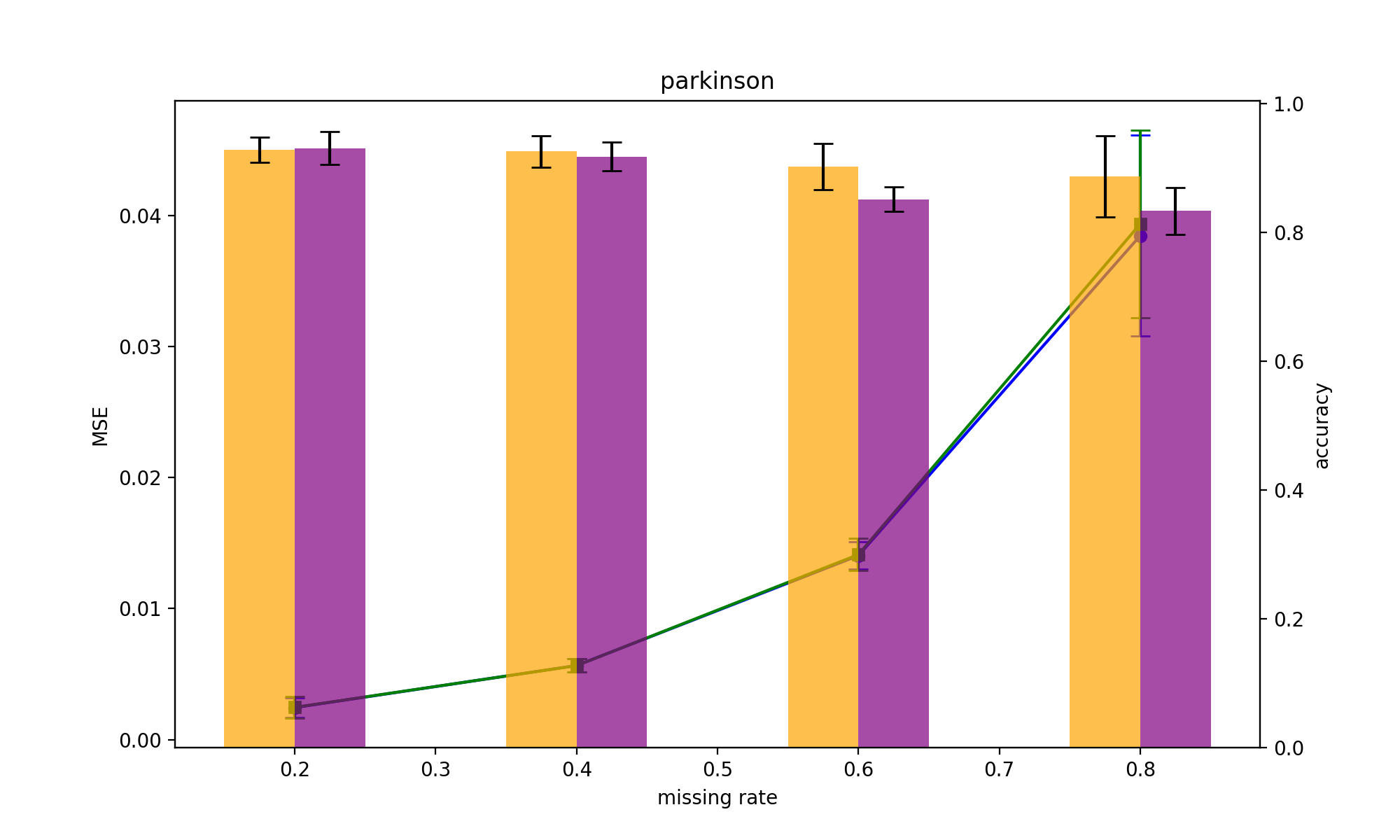}
        \label{fig-parkinson-mice}
    \end{minipage}
    \begin{minipage}[b]{0.35\textwidth}
        \centering
        \includegraphics[width = 1.05\textwidth]{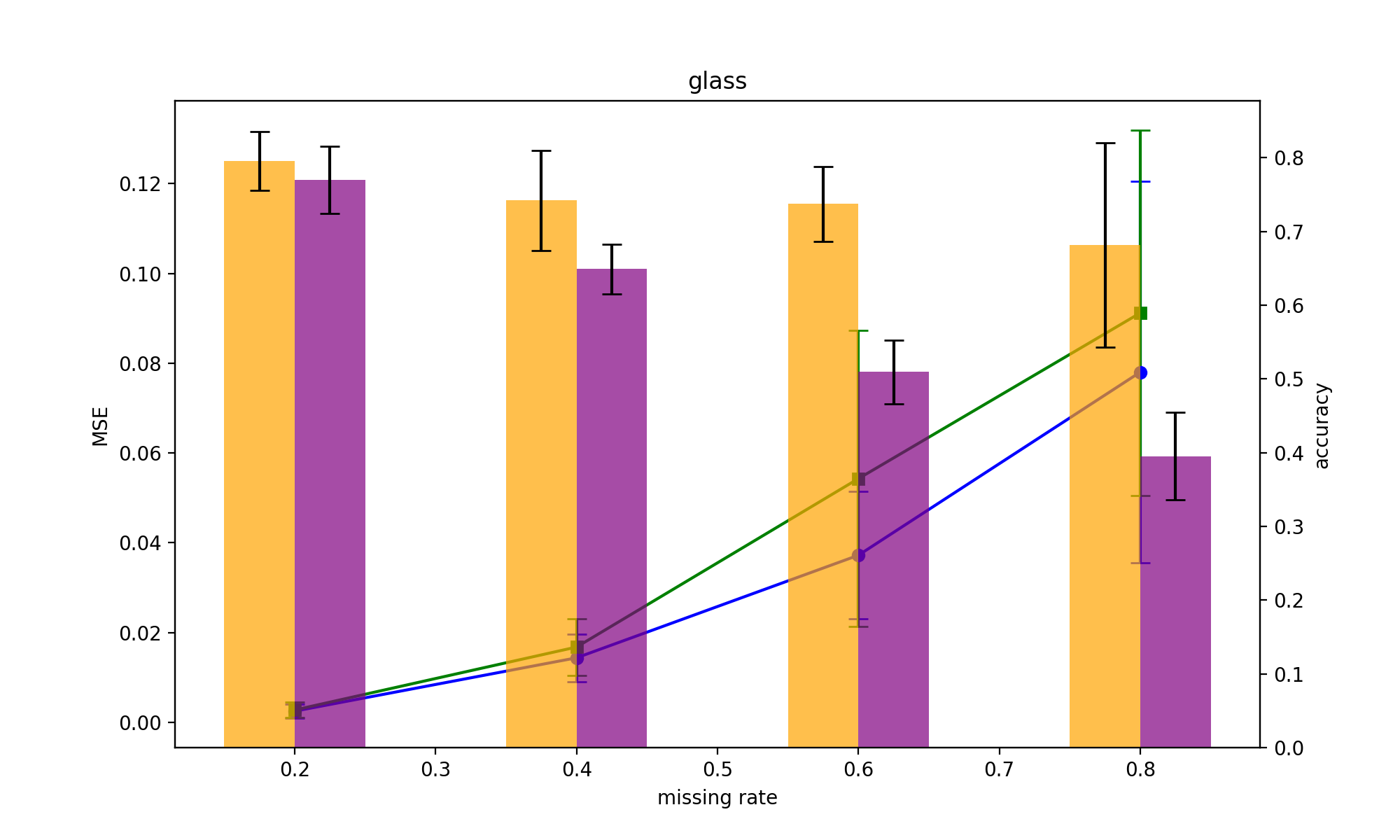}
        \label{fig-glass-mice}
    \end{minipage}
    \caption{Performance of IUL compared to DI with MICE for classification tasks.}
    \label{fig-iul-di-mice}
\end{figure*}

\begin{figure*}[!ht]
    \centering
    \includegraphics[width = 0.35\textwidth]{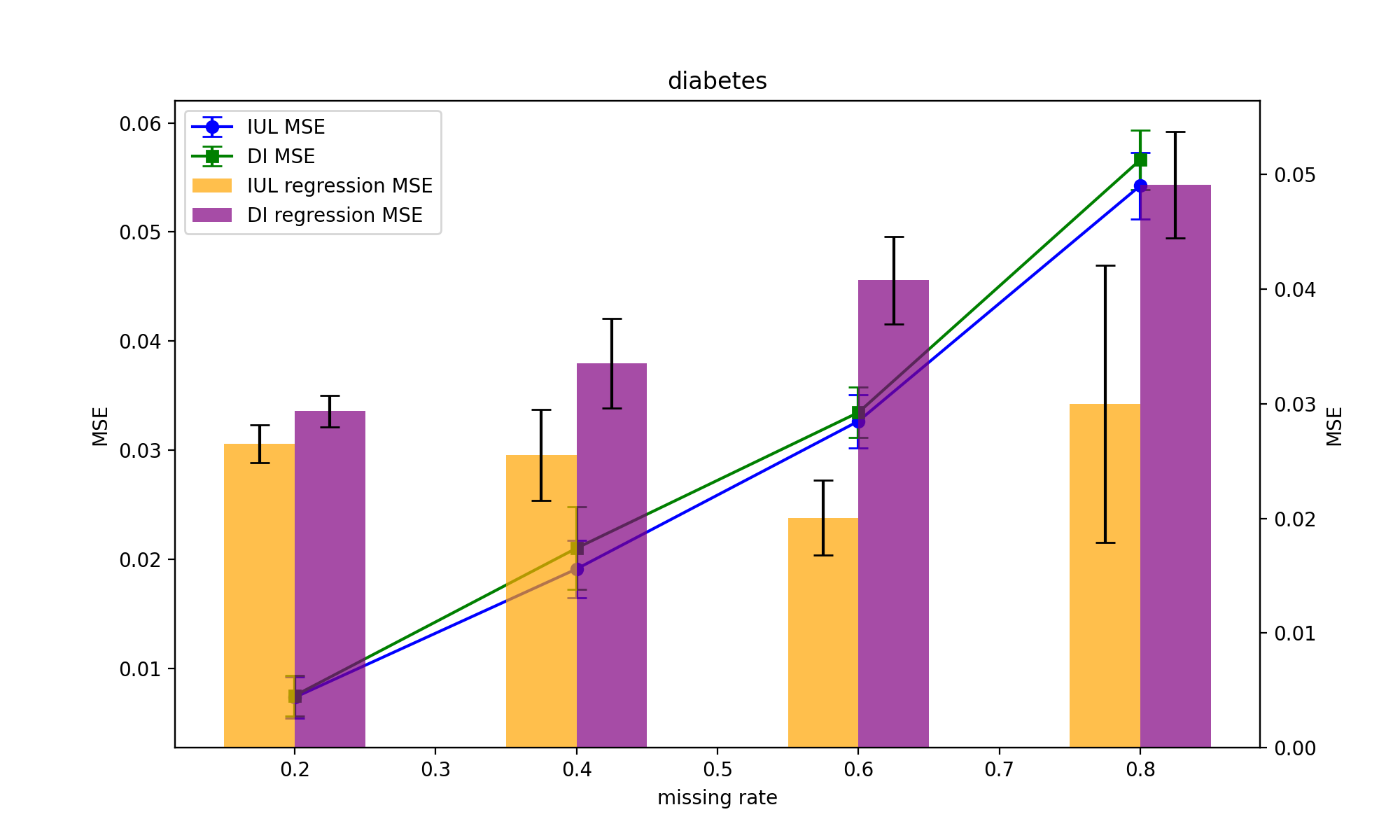}
    \caption{Performance of IUL compared to DI  with MICE for regression tasks.}
    \label{fig-iul-di-diabetes-regression}
\end{figure*}
The results of the experiments on the MSE and accuracy are shown in Figure~\ref{fig-iul-di}, \ref{fig-iul-di-regression}, \ref{fig-iul-di-mice}, and \ref{fig-iul-di-diabetes-regression}. Since Soybean and Car datasets contain only categorical features, the plot of these datasets for this regression task is not available. In Figure~\ref{fig-iul-di} and \ref{fig-iul-di-mice}, accuracy results are not provided for the California and diabetes datasets due to their regression nature instead of classification tasks, therefore, we introduce regression MSE for comparison in Figure~\ref{fig-iul-di-regression} and \ref{fig-iul-di-diabetes-regression}. Also, note that the result of the implementation on the California dataset is not available in Figure \ref{fig-iul-di-diabetes-regression} due to insufficient memory.

Considering the performance on the imputation task across all figures, IUL consistently achieves a lower MSE than DI, particularly as the missing rate increases. This highlights IUL's robustness in handling missing data. For instance, on the Iris dataset (Figure~\ref{fig-iul-di}), the difference in MSE between IUL and DI is minimal at a missing rate of 20\%, but at 80\% missing data, IUL achieves an MSE less than half that of DI (0.03 compared to 0.08).

When analyzing accuracy in Figures~\ref{fig-iul-di} and \ref{fig-iul-di-mice}, a consistent pattern emerges where IUL outperforms DI in classification tasks. This improvement becomes more pronounced as the missing rate increases, indicating that IUL not only reduces error in terms of MSE but also enhances classification performance. Regarding regression MSE, as shown in Figures~\ref{fig-iul-di-regression} and \ref{fig-iul-di-diabetes-regression}, IUL demonstrates superior performance compared to DI. For example, in the Diabetes dataset, IUL maintains a consistently low regression MSE in the range of [0.02, 0.03] across both figures, while DI's regression MSE steadily rises from approximately 0.03 to 0.05.

\section{Conclusion and future works}\label{sec:conclusion}
In conclusion, this study has presented two novel strategies, \textit{CBMI} and \textit{IUL}, that share a common foundation in stacking input and output. These algorithms leverage the often overlooked labels in the training data, significantly improving imputation/classification quality. Specifically, CBMI, which integrates the missForest algorithm, enables simultaneous imputation of the training and testing label and input and is capable of handling data training with missing labels without prior imputation. Both CBMI and IUL have demonstrated their versatility by applying to continuous, categorical, or mixed-type data. Experimental results have shown promising improvements in mean squared error or classification accuracy, especially in dealing with imbalanced data, categorical data, and small sample data, indicating their potential to enhance data analysis and model performance.  
A remaining limitation of CBMI and IUL, however, is that they are primarily applicable to scenarios where all data is collected in advance, and no new data is expected. 

Also, in the future, we would like to look at the performance of various imputation techniques other than missForest and MICE by using the imputed data for a downstream predictive task and compare its performance with models trained on data imputed using other methods.
Another area for improvement of this work is that it is not built for MNAR data,  where the missingness depends on unobserved factors. So, in the future, we want to extend CBMI and IUL to handle this scenario. 

Also, we want to explore CBMI and IUL's capability to handle input with missing labels further and compare them to clustering algorithms. In addition, we also want to study the application of CBMI in semi-supervised learning, as well as its performance compared to the pseudo-label approach as in \cite{lee2013pseudo}. Specifically, in semi-supervised learning, there are samples without labels. So, we can regard these samples as samples with missing labels. Moreover, we will investigate if similar strategies work for regression. Moreover, when using CBMI with missForest, an algorithm that is built upon Random Forests, it is also likely that CBMI also inherits the properties of Random Forest, such as robustness to noise, the ability to handle outliers, and multicollinearity. Therefore, in the future, we will conduct experiments to confirm whether CBMI also inherits these qualities.
\vspace{-5mm}
\appendix
\section{Proof of theorem \ref{theo-1}}\label{appendix-proof}
\textbf{Theorem 1.} Assume that we have a data $\mathcal{D}=(\mathbf{x},\mathbf{z},\mathbf{y})$ of $n$ samples. Here, $\mathbf{x}$ contains missing values, $\mathbf{z}$ is fully observed and $\mathbf{y}$ is a label feature. For MICE imputation, with the IUL strategy, we construct the model $\hat{x}=\hat{\gamma}_o+\hat{\gamma}_1z+\hat{\gamma}_2y.$
Meanwhile, DI ignores label feature $\mathbf{y}$ and use the model 
$\hat{x}'=\hat{\gamma}_o+\hat{\gamma}_1z.$
Without loss of generality, assume that $y_i \ge 0, i=1,2,...,n$ for label encoding. Let denote 
\vspace{-2mm}
\begin{align}\label{Eq_Ei}
    \mathcal{E}_i &= \hat{\gamma}_2^2(y_i-2\mu_{\mathbf{y}})+2\hat{\gamma}_1\hat{\gamma}_2(z_i-\mu_{\mathbf{z}}), \quad i=1,2,\dots,n.
\end{align}
Here, the value of $\mathcal{E}_i$ could be negative or non-negative. Thus, we distinguish between two parts by denoting
\begin{align}
    \mathcal{V}^+ = \sum_{i\in \mathcal{I}^+}y_i\mathcal{E}_i, \text{ where }\mathcal{I}^+ &= \{i : \mathcal{E}_i \ge 0 \}, \label{Eq_V+}\\
    \mathcal{V}^- = \sum_{i\in \mathcal{I}^-}y_i\mathcal{E}_i, \text{ where }  \mathcal{I}^- &= \{i : \mathcal{E}_i < 0 \}. \label{Eq_V-}
\end{align}
Then, IUL outperforms DI (i.e. $SSE_{IUL}\le SSE_{DI}$) if and only if $ \mathcal{V}^+ + \mathcal{V}^- \ge 0,$ where $SSE_{IUL}, SSE_{DI}$ are the Sum of Square Error for the model based on IUL and DI, respectively.
\\
\textbf{Proof.}
We first recall in linear regression, $SSE = SST - SSR$ where $SST$ is the Sum of Squares Total and $SSR$ is the Sum of Squares due to Regression. For each model relies on IUL and DI we have
\vspace{-5mm}
\begin{align*}
    SSE_{IUL} &= \sum_{i=1}^{n}(\hat{x}_i-x_i)^2 =\sum_{i=1}^{n}(x_i - \mu_{\mathbf{x}})^2 - \sum_{i=1}^{n}(\hat{x}_i-\mu_{\mathbf{x}})^2, \\
    SSE_{DI} &= \sum_{i=1}^{n}(\hat{x}'_i-x_i)^2 =\sum_{i=1}^{n}(x_i - \mu_{\mathbf{x}})^2 - \sum_{i=1}^{n}(\hat{x}'_i-\mu_{\mathbf{x}})^2.
\end{align*}
Then, the difference between the Sum of Square Error for the model based on DI and IUL is given
\begin{align*}
    SSE_{DI} - &SSE_{IUL} =\sum_{i=1}^{n}(\hat{x}_i-\mu_{\mathbf{x}})^2 - \sum_{i=1}^{n}(\hat{x}'_i-\mu_{\mathbf{x}})^2\\
    &= \sum_{i=1}^{n}(\hat{x}_i-\hat{x}'_i +\hat{x}'_i - \mu_{\mathbf{x}})^2 - \sum_{i=1}^{n}(\hat{x}'_i-\mu_{\mathbf{x}})^2\\
    &= \sum_{i=1}^{n}(\hat{x}_i-\hat{
    x}'_i)^2 + 2\sum_{i=1}^{n}(\hat{x}_i-\hat{
    x}'_i)(\hat{x}'_i-\mu_{\mathbf{x}})\\
    &=\sum_{i=1}^{n} (\hat{\gamma}_2y_i)^2+2\sum_{i=1}^{n}\hat{\gamma}_2y_i(\hat{\gamma}_1z_i+\hat{\gamma}_o-\mu_{\mathbf{x}})   \\
    &=\sum_{i=1}^{n}\hat{\gamma}_2y_i(\hat{\gamma}_2y_i+2(\hat{\gamma}_1z_i+\hat{\gamma}_o-\mu_{\mathbf{x}}))\\
    &=\sum_{i=1}^{n}\hat{\gamma}_2y_i(\hat{\gamma}_2y_i+2(\hat{\gamma}_1z_i-\hat{\gamma}_1\mu_{\mathbf{z}} - \hat{\gamma}_2\mu_{\mathbf{y}}))\\
    &=\sum_{i=1}^{n}y_i(\hat{\gamma}_2^2(y_i-2\mu_{\mathbf{y}})+2\hat{\gamma}_1\hat{\gamma}_2(z_i-\mu_{\mathbf{z}}))
\end{align*}
From the equation~\eqref{Eq_Ei} and also by the notation~\eqref{Eq_V+},~\eqref{Eq_V-}, we can imply $ SSE_{DI} - SSE_{IUL}= \sum_{i=1}^n y_i\mathcal{E}_i = \sum_{i\in \mathcal{I}^+}y_i\mathcal{E}_i + \sum_{i\in \mathcal{I}^-}y_i\mathcal{E}_i  = \mathcal{V}^+ + \mathcal{V}^- $
Therefore, $SSE_{DI}\ge SSE_{IUL}$ if and only if $ \mathcal{V}^+ + \mathcal{V}^- \ge 0$. In other words, IUL outperforms DI if and only if $ \mathcal{V}^+ + \mathcal{V}^- \ge 0$.

\bibliographystyle{IEEEtran}
\bibliography{ref}

\end{document}